\documentclass[letterpaper, 10 pt, conference]{ieeeconf}  

\IEEEoverridecommandlockouts                              

\usepackage{amsmath,amsfonts}
\usepackage{algorithmic}
\usepackage{algorithm}
\usepackage{array}
\usepackage[caption=false,font=normalsize,labelfont=sf,textfont=sf]{subfig}
\usepackage{textcomp}
\usepackage{stfloats}
\usepackage{url}
\usepackage{verbatim}
\usepackage{graphicx}
\usepackage{cite}
\usepackage{multirow} 
\usepackage{booktabs} 
\usepackage{caption}
\usepackage{balance}
\usepackage{mathtools}
\usepackage{tabularx}

\title{\LARGE \bf
Flying to Image-Specified Objects: 3D Quadrotor Navigation via Cross-Graph Memory and Viewpoint Planning
}

\author{Junjie Gao$^{*}$, Yuqi Chen$^{*}$, Yongzhou Pan, Yaosheng Deng, Jiaping Xiao, and Mir Feroskhan$^{\dagger}$
\thanks{$^{*}$Junjie Gao and Yuqi Chen contributed equally to this work.}
\thanks{$^{\dagger}$Mir Feroskhan is the corresponding author.}
\thanks{Junjie Gao, Yuqi Chen, Yongzhou Pan, Yaosheng Deng, Jiaping Xiao, and Mir Feroskhan are with Nanyang Technological University, Singapore.}%
}

\begin{document}

\maketitle
\thispagestyle{empty}
\pagestyle{empty}

\begin{abstract}
Instance-Specific Image-Goal Navigation (InstanceImageNav) requires a robot to navigate toward the exact object instance depicted in a query image. Extending this task to quadrotors is challenging due to continuous 3D control, limited field of view (FOV), and safety constraints, which make successful navigation highly dependent on selecting informative viewpoints. 
We propose a hierarchical navigation framework for quadrotor InstanceImageNav that separates high-level decision making from low-level motion execution. Instead of navigating directly to spatial locations, the system generates viewpoint-aware action nodes around frontier regions and potential target objects, enabling the robot to explore while maintaining informative viewpoints for detecting the target instance. A lightweight semantic memory maintains object-level and observation-level context, allowing semantic cues to propagate to candidate action nodes for decision making.
A learning-based policy selects the most promising action node, and a trajectory planner generates dynamically feasible 3D flight paths for safe execution. Experiments in simulation demonstrate consistent improvements over strong baselines, and real-world quadrotor flights validate the practicality and robustness of the proposed framework.
\end{abstract}


\section{INTRODUCTION}

Vision-based target-driven navigation has become an important capability for autonomous robots in applications such as household service \cite{ref1}, infrastructure inspection \cite{ref2}, and search-and-rescue operations \cite{ref3}. In this setting, a robot must navigate toward a target specified through visual information rather than explicit coordinates. A particularly challenging formulation is \emph{Instance-Specific Image-Goal Navigation} (InstanceImageNav), where the robot must locate the exact object instance depicted in a query image \cite{ref4}. Unlike ObjectNav, where any instance of a category is acceptable, InstanceImageNav requires identifying the correct object instance under significant appearance and viewpoint variations.

Extending InstanceImageNav to quadrotor platforms introduces additional challenges. Quadrotors operate in fully three-dimensional environments with continuous control and typically rely on forward-facing cameras with limited field of view (FOV). In contrast, many existing navigation approaches assume planar motion~\cite{ref5,ref6,ref7,ref10} or rely on panoramic observations to reduce viewpoint ambiguity~\cite{ref8,ref9}. Under limited FOV sensing, the visibility of the target object strongly depends on the robot’s viewpoint, making navigation not only a path planning problem but also a viewpoint selection problem. End-to-end policies that directly map observations to control commands often struggle in such settings due to the large action space and the difficulty of learning flight dynamics in continuous 3D environments. Moreover, these policies are typically sensitive to domain shifts, which widens the sim-to-real gap in real-world deployments~\cite{ref11}.

Modular navigation frameworks offer a promising alternative by separating high-level decision making from low-level motion execution~\cite{ref5,ref9,ref12,ref42}. However, the effectiveness of such systems largely depends on how candidate navigation targets are generated and selected. Existing approaches often treat frontiers or predicted waypoints as navigation targets, which may not provide informative viewpoints for detecting visually specified objects under limited FOV. As a result, robots may explore large areas without observing the target instance even when it is nearby.

To address these challenges, we propose a hierarchical navigation framework that explicitly reasons about candidate viewpoints during navigation. Instead of navigating directly to frontiers or spatial locations, our method generates viewpoint-aware action nodes that consider both camera visibility and quadrotor motion constraints. A lightweight semantic memory maintains object-level and observation-level information, allowing the policy to leverage accumulated scene context when selecting action nodes. The selected node is then executed by a trajectory planner that produces a dynamically feasible flight path. This design integrates semantic reasoning, viewpoint-aware exploration, and classical trajectory planning into a unified framework for quadrotor InstanceImageNav.

The main contributions of this work are summarized as follows:
\begin{itemize}
    \item We formulate the InstanceImageNav problem for quadrotor platforms operating in continuous 3D environments with limited FOV sensing, extending visually specified navigation to aerial robots with full 3D motion and viewpoint constraints.

    \item We propose a viewpoint-aware action node generation and selection strategy that explicitly accounts for camera visibility and quadrotor motion constraints, enabling efficient exploration and reliable target observation.

    \item We develop a hierarchical navigation framework that integrates semantic memory, learning-based decision making, and trajectory planning for safe quadrotor navigation, and validate the approach through extensive simulation and real-world experiments.
\end{itemize}

\begin{figure*}[t] 
    \centering
    \includegraphics[width=\textwidth]{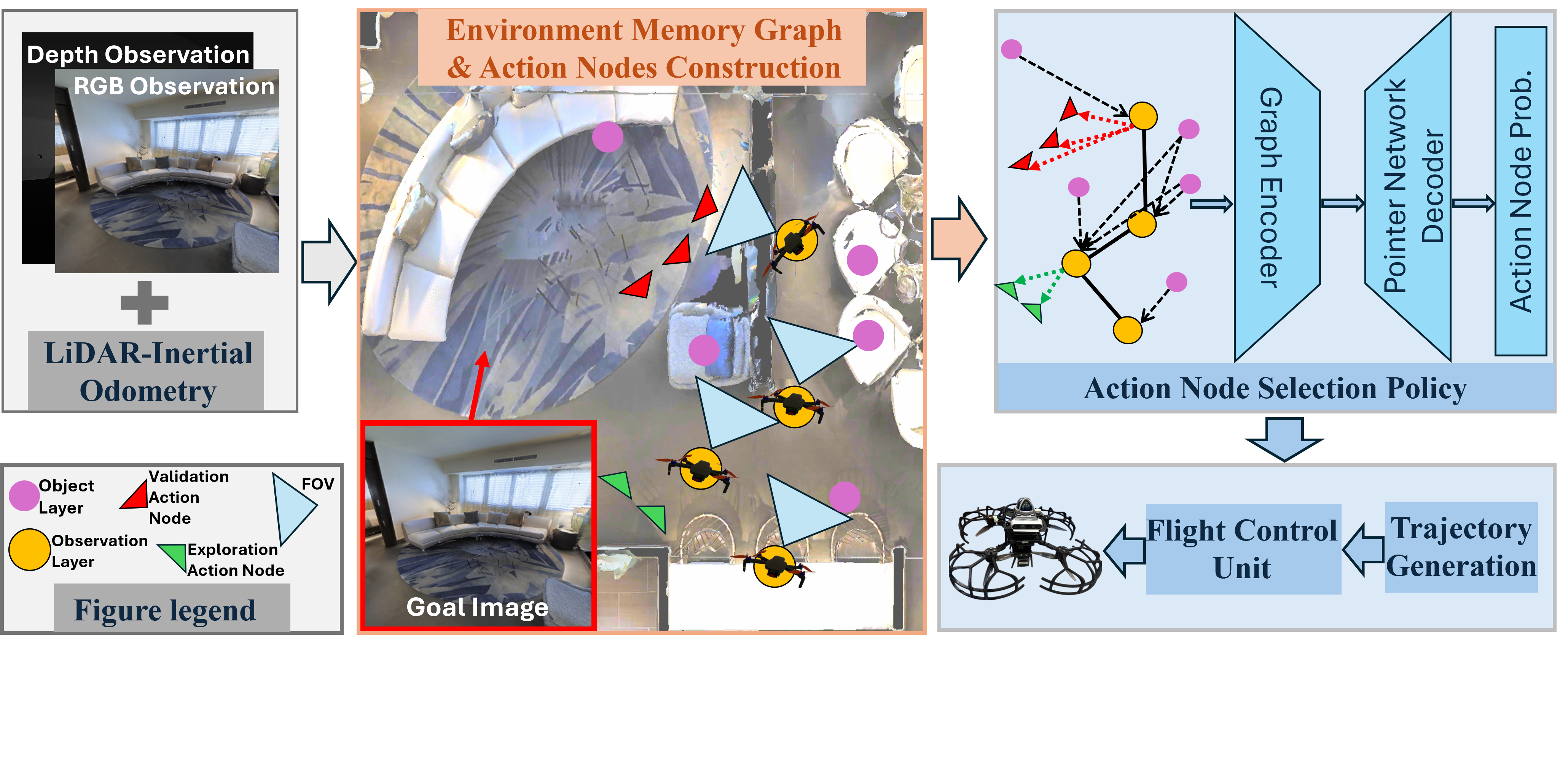}
    \caption{Overall framework of the proposed quadrotor InstanceImageNav framework. RGB-D observations and the goal image are used to build an environment memory graph with object and observation layers. A graph encoder and pointer-network decoder predict selection probabilities for candidate action nodes, which then guide trajectory generation and flight control toward the goal.}
    \label{fig:scene_example}
\end{figure*}

\section{Related Works}
\subsection{Visual Navigation}
Existing visual navigation methods can be broadly categorized into end-to-end approaches and modular solutions \cite{ref44}. End-to-end navigation methods directly map ego-centric visual inputs into actions \cite{ref6, ref7, ref8, ref14}. These approaches address challenges such as visual scene understanding and large environment exploration by incorporating careful reward shaping, auxiliary tasks, and advanced memory modules. For example, Maksymets \emph{et al.} alleviate the overfitting problem by designing the ExploreTillSeen reward, a phased reward mechanism that separates exploration and navigation \cite{ref20}. Bono \emph{et al.} introduce two pretext tasks of cross-view completion and relative pose and visibility estimation to solve the wide-baseline visual correspondence problem \cite{ref21}. Although end-to-end methods achieve promising performance in simulators, their effectiveness significantly degrades when applied to real-world scenarios. To address this limitation, modular methods decompose the task into distinct components. Ramakrishnan \emph{et al.} introduced a potential field function to address the exploration challenges in Object Goal Navigation \cite{ref23}. The framework proposed in \cite{ref5} explicitly switches between exploration, verification, and exploitation stages, enabling the robot to make contextually appropriate decisions. However, this strict separation prevents the exploration policy from learning to approach the goal effectively. Ye \emph{et al.} propose a hierarchical policy architecture: The policy network determines a local goal, followed by a sampling-based planning module. However, they adopt a panoramic camera and a 2D experimental setting ~\cite{ref9}.

\subsection{Environment representation}
In visual navigation tasks, environment representation functions as the robot's memory. There are three main types of explicit representations: metric graphs \cite{ref24, ref25}, topological graphs \cite{ref8, ref13, ref16, ref17}, and their combinations \cite{ref9, ref12}. Active Neural SLAM \cite{ref24} constructs a top-down 2D metric map to assist global goal estimation for the navigation policy. Exp4Nav \cite{ref25} builds both the global and the egocentric metric map, which are cropped and embedded for action generation. For topological memory, work in \cite{ref15} forms a topological graph through exploration before training. Then, the robot navigates to a graph node based on the trained policy network. Shah \emph{et al.} build a topological graph and use the Dijkstra algorithm to perform the navigation task \cite{ref16}. Each graph node corresponds to one of the previous observations, and the edge weights are determined by a learned distance function. Kim \emph{et al.} construct two graphs representing observations and detected objects \cite{ref8}. They further design a mixing mechanism to exchange nodes between these graphs, enabling better integration of their respective node features. The third type of method employs both metric maps and sparse topological graphs. Ravichandran \emph{et al.} design the navigation policy based on the pre-constructed 3D scene graphs, which represent the environment in five layers from fine to coarse \cite{ref26}. Cui \emph{et al.} incrementally build a topological graph for action node selection while maintaining a local metric map for obstacle avoidance \cite{ref9}. To facilitate decision-making and enable 3D planning, we adopt a hybrid representation and design a cross-graph information-passing mechanism that injects semantic cues into action nodes derived from the metric map. 

\section{Problem Definition}

The quadrotor InstanceImageNav task is defined as a vision-based target-driven navigation problem in which the quadrotor must locate and approach a specific object instance $O$ depicted in a query RGB image $I_g$. The query image specifies the visual appearance of the target instance but does not provide its spatial location.
The quadrotor is initialized at a random pose $x_0 = (p_0, q_0)$, where $p_0 \in \mathbb{R}^3$ denotes the 3D position and $q_0 \in \mathbb{R}^4$ represents the orientation. At each time step $t$, the quadrotor receives egocentric RGB-D observations $o_t = (o_t^{\text{RGB}}, o_t^{\text{D}})$, where $o_t^{\text{RGB}} \in \mathbb{R}^{3 \times H \times W}$ is the RGB image and $o_t^{\text{D}} \in \mathbb{R}^{H \times W}$ is the depth map. The quadrotor also has access to its current pose $x_t = (p_t, q_t)$.
The objective is to navigate toward the target object instance $O$. An episode is considered successful when the quadrotor reaches a position sufficiently close to the target instance and the object is visible in the current observation. Formally, success is achieved when $\|p_t - p_O\|_2 \leq \epsilon$
and the object instance $O$ appears in the RGB observation $o_t^{\text{RGB}}$.

\section{Environment Memory and Action Node Generation}

Our navigation framework relies on two key components: an environment memory that stores accumulated semantic and spatial information, and a set of action nodes that represent candidate viewpoints for future navigation. 
Unlike conventional waypoint-based navigation where nodes correspond to intermediate spatial locations along a path, an action node represents a viewpoint from which the quadrotor can observe informative regions or inspect potential target objects. 
This design is particularly important for InstanceImageNav under limited FOV sensing, where selecting informative viewpoints is crucial for discovering the target instance.

\subsection{Environment Memory Graph Update}

To maintain an interpretable representation of the explored environment, we construct a heterogeneous memory graph composed of two layers: an object layer and an observation layer.

\textbf{Object layer.}
Each node in the object layer represents a detected object instance and is characterized by an attribute vector
\begin{equation}
f^{obj}_t = (r^{obj}_t, r^{obj}_g, c^{obj}_t, c^{obj}_g, b^{obj}_t),
\end{equation}
where \(r^{obj}_t\) and \(r^{obj}_g\) denote the visual features of the observed object and the target object, respectively, \(c^{obj}_t\) and \(c^{obj}_g\) represent the corresponding semantic embeddings, and \(b^{obj}_t\) denotes the estimated 3D bounding box size.

We employ YOLOE~\cite{ref40} as the object detector due to its open-vocabulary capability, which provides richer semantic information for navigation. The semantic feature \(c^{obj}_t\) is obtained by encoding the detected category label using the CLIP text encoder~\cite{ref41}. The visual feature \(r^{obj}_t\) is extracted by pooling backbone features within the detected region, capturing the object's visual appearance. The geometric attribute \(b^{obj}_t\) is estimated by back-projecting pixels inside the 2D bounding box into 3D space using the depth observation.
To avoid redundant nodes, newly detected objects are compared with existing nodes based on their geometric proximity and visual similarity. If a corresponding node already exists, its features are updated using the detection with the higher confidence score.

\textbf{Observation layer.}
The observation layer stores visual context associated with visited viewpoints. Each observation node is represented by a feature vector
\begin{equation}
f^{obs}_t = (r^{obs}_t, g^{obs}_t, c^{obs}_t),
\end{equation}
where \(r^{obs}_t\) denotes the visual embedding extracted from the RGB observation, \(g^{obs}_t\) denotes the visual feature of the goal image, and \(c^{obs}_t\) represents the spatial position of the node.

To construct observation features, we employ a pretrained ResNet-18 network to extract visual embeddings from \(o_t^{RGB}\). To prevent storing redundant observations, a pretrained similarity encoder~\cite{ref39} is used to measure the similarity between the current observation and previously stored nodes. If the similarity score falls below a predefined threshold, a new observation node is inserted into the graph and connected to the previous node in temporal order.

Edges between the object and observation layers associate detected objects with the viewpoints from which they were observed. The edge feature is defined as the Euclidean distance between the object location and the corresponding observation node.

\subsection{Action Node Set Update}

The action node set contains candidate viewpoints that the quadrotor may visit next. These nodes are generated to support two complementary behaviors: exploration of unseen regions and inspection of potential target objects.

\textbf{Exploration action nodes.}
To explore unexplored areas of the environment, we maintain a 3D occupancy grid map for frontier extraction. At each step \(t\), frontier voxels are first identified and grouped using Euclidean clustering. To ensure manageable cluster sizes, large clusters are further split into smaller regions.

Around each frontier cluster, candidate viewpoints are sampled in a cylindrical coordinate system~\cite{ref19}. Each viewpoint \(V_i\) is evaluated using the following scoring function:
\begin{equation}
\mathcal{S}(V_i) =
\lambda_{\mathrm{IG}} \cdot IG_i +
\lambda_{\mathrm{yaw}} \cdot |\Delta \psi_i| -
\lambda_{\mathrm{path}} \cdot D_i,
\end{equation}
where \(IG_i\) denotes the number of unknown voxels visible from viewpoint \(V_i\), \(D_i\) denotes the path length from the current pose to \(V_i\), and \(|\Delta \psi_i|\) represents the yaw change, which encourages diverse viewing directions. Candidate viewpoints are greedily selected in descending order of \(\mathcal{S}(V_i)\) until the frontier region is sufficiently covered.

\textbf{Shortcut action nodes.}
To accelerate task execution, shortcut nodes are generated when the current observation exhibits strong visual similarity to the goal image. Given the goal image \(I_g\) and the current RGB observation \(o_t^{RGB}\), we extract keypoints and descriptors using XFeat~\cite{ref30}:
\begin{equation}
(K_g, D_g) = \text{XFeat}(I_g), \quad
(K_t, D_t) = \text{XFeat}(o_t^{RGB}),
\end{equation}
and perform feature matching using LightGlue~\cite{ref31}. If the number of matched keypoints exceeds a predefined threshold, the system retrieves the object layer node associated with the current observation node whose visual feature \(r^{obj}_t\) exhibits the highest similarity to the target feature \(r^{obj}_g\).

Around the matched object node, candidate viewpoints are sampled using the same cylindrical sampling strategy. These viewpoints are evaluated according to
\begin{equation}
\mathcal{S}(V_i)=
\lambda_{\mathrm{cov}} \cdot Cov_i +
\lambda_{\mathrm{yaw}} \cdot |\Delta \psi_i| -
\lambda_{\mathrm{path}} \cdot D_i,
\end{equation}
where \(Cov_i \in [0,1]\) denotes the fraction of the object’s 3D bounding box surface visible from viewpoint \(V_i\). This scoring encourages viewpoints that provide better visibility of the suspected target object.
Importantly, the quadrotor is not commanded to directly approach the detected object location once a match is found. Instead, viewpoint-aware action nodes are generated to ensure that the object can be inspected from safe and informative viewing angles.
After generation, each action node is connected to its nearest observation node, where the edge weight is defined as the safe A* path length between them. This connection allows the navigation policy to leverage the accumulated semantic and spatial context stored in the environment memory when selecting the next action node.

\section{Hierarchical Navigation Pipeline}

We adopt a hierarchical navigation pipeline that decouples high-level decision making from low-level motion execution. 
At the high level, a learning-based policy selects the most promising action node by leveraging semantic and visual context stored in the environment memory graph. 
At the low level, an optimization-based planner computes a dynamically feasible and collision-free trajectory toward the selected node.

\subsection{Action Node Selection Policy}

The action node selection problem is formulated as a Partially Observable Markov Decision Process (POMDP) defined by the tuple 
$(S, A, P, R, \Omega, \mathcal{O}, \gamma)$.
The action space \(A\) corresponds to the set of available action nodes. 
The observation \(\Omega\) consists of the quadrotor pose \(x_t = (p_t, q_t)\) together with the environment memory graph constructed up to time step \(t\). 
The reward function is defined as

\begin{equation}
\label{eq2}
r(t)=r_t^g + r_t^d + r_t^{s},
\end{equation}

where \(r_t^g\) is a sparse goal reward provided only upon task completion. 
The dense reward term is defined as 
\(r_t^d=\omega_d (d_{t-1}-d_t)\), where \(d_t\) denotes the A* path length from the current pose to the target position. 
A slack penalty \(r_t^{s}=-\omega_s\) is applied at every step to encourage shorter trajectories.

\textbf{Policy architecture.}
The policy network consists of a graph encoder followed by a pointer-network decoder and a value head.
The encoder extracts contextualized features from the environment memory graph, while the decoder produces a probability distribution over candidate action nodes.
We train the policy using Proximal Policy Optimization (PPO)~\cite{ref32}.

\textbf{Encoder.}
The encoder contains \(L\) stacked layers. 
Each layer first updates observation-layer nodes using graph attention, then injects semantic information from the object layer, and finally projects the updated features to the associated action nodes.

For the \(l\)-th layer, the observation node features are first updated using graph attention:

\begin{equation}
\label{eq:obs_gat}
\mathbf{h}_{i,obs}^{l}{}' =
\sigma \!\left(
\sum_{k \in \mathcal{N}_{obs}(i)}
\alpha_{ik}^{(l)} \mathbf{W}_{obs}^{(l)} \mathbf{h}_{k,obs}^{l}
\right),
\end{equation}

where \(\mathcal{N}_{obs}(i)\) denotes the neighbors of observation node \(i\), 
and \(\alpha_{ik}^{(l)}\) are attention coefficients computed following~\cite{ref33}.

Next, semantic context is propagated from the object layer through cross-layer message passing:

\begin{equation}
\label{eq:obj2obs}
\mathbf{h}_{i,obs}^{l}{}'' =
\sigma \!\left(
\sum_{m \in \mathcal{N}_{obj-obs}(i)}
\alpha_{im}^{cross} \mathbf{W}_{cross}^{(l)} \mathbf{h}_{m,obj}^{l}
\right),
\end{equation}

where \(\mathcal{N}_{obj-obs}(i)\) denotes the object nodes connected to observation node \(i\).

The observation node representation is then updated by combining spatial and semantic information:

\begin{equation}
\label{eq:obs_update}
\mathbf{h}_{i,obs}^{l+1} =
MLP_{obs}\!\Big([\mathbf{h}_{i,obs}^{l}{}', \mathbf{h}_{i,obs}^{l}{}'']\Big).
\end{equation}

Finally, the updated observation features are projected to the associated action nodes:

\begin{equation}
\label{eq:cross_graph}
\mathbf{h}_{j,c}^{l+1} =
\mathbf{W}_{c}^{(l)} \mathbf{h}_{i,obs}^{l+1},
\end{equation}

where \(i\) denotes the observation node connected to action node \(j\). 
This mapping transfers the accumulated spatial and semantic context to the action node representation.

\textbf{Decoder.}
The decoder adopts a pointer-network architecture~\cite{ref34} to select an action node.
The robot state feature \(x_t\) and the goal image feature \(f_g\) are first encoded by MLP layers and concatenated to form a query vector \(q_c\).
The query attends over the encoded action node features \(\mathbf{h}_{c}^{L+1}\), producing attention scores that are normalized into weights \(\alpha_c\).
These weights define the stochastic policy
\(\pi(v_i | o_t) = \alpha_{ci}\).

A value head shares the encoder backbone and predicts the state value through an independent MLP, enabling PPO optimization.

\subsection{Trajectory Generation}

Once an action node is selected, a dynamically feasible trajectory is generated toward the corresponding viewpoint.
Following~\cite{ref19}, we employ a two-stage planning scheme consisting of hybrid A* search and cubic B-spline trajectory optimization.

Hybrid A* first generates a feasible initial path. 
This path is then refined by solving

\begin{equation}
\label{eq:traj_opt}
\min_{x(t)} J =
\begin{aligned}[t]
&\lambda_j \!\!\int_0^{T} \|\dddot{x}(t)\|^2 \, dt \\
&\mathllap{+}\lambda_v \!\!\int_0^{T} \phi(\|\dot{x}(t)\|-v_{\max}) \, dt \\
&\mathllap{+}\lambda_a \!\!\int_0^{T} \phi(\|\ddot{x}(t)\|-a_{\max}) \, dt
\end{aligned}
\end{equation}
where \(x(t)\) denotes the position trajectory, 
\(\dot{x}(t)\) and \(\ddot{x}(t)\) represent velocity and acceleration, respectively. 
The first term minimizes jerk to ensure smooth motion, while the remaining terms softly enforce velocity and acceleration constraints using the hinge loss \(\phi(z)=\max(0,z)^2\).
Finally, the yaw profile is independently generated by interpolating from the current yaw to the desired yaw of the selected action node, ensuring that the quadrotor reaches the viewpoint with the intended viewing direction.

\begin{table*}[t]
\renewcommand{\arraystretch}{1.1}
\centering
\caption{Comparative study results.}
\label{tab:baseline}
\small
\begin{tabularx}{0.95\textwidth}{>{\raggedright\arraybackslash}p{4.2cm}*{9}{>{\centering\arraybackslash}X}}
\toprule
\multirow{2}{*}{\textbf{Method}}
 & \multicolumn{3}{c}{Easy} & \multicolumn{3}{c}{Medium} & \multicolumn{3}{c}{Hard} \\
\cmidrule(lr){2-10}
 & SR & SPL & CFR & SR & SPL & CFR & SR & SPL & CFR \\
\midrule
OVRL-v2-IIN \cite{ref14}          & 0.47 & 0.23 & 0.67 & 0.24 & 0.11 & 0.55 & 0.09 & 0.03 & 0.83 \\
FUEL \cite{ref19}          & 0.85 & 0.43 & 0.20 & 0.53 & 0.16 & 0.28 & 0.39 & 0.05 & 0.44 \\
MOD-IIN \cite{ref42}          & 0.71 & 0.33 & 0.27 & 0.59 & 0.21 & 0.35 & 0.21 & 0.09 & 0.39 \\
Modular ImageNav \cite{ref12}      & 0.73 & 0.48 & 0.26 & 0.42 & 0.33 & 0.31 & 0.40 & 0.18 & 0.38 \\
IEVE \cite{ref5}   & 0.85 & 0.53 & 0.36 & 0.51 & 0.29 & 0.46 & 0.44 & 0.21 & 0.45 \\
Topo-Metric ImageNav \cite{ref9}   & \textbf{0.89} & 0.50 & 0.30 & 0.57 & 0.34 & 0.46 & 0.46 & 0.29 & 0.43 \\
\midrule
{Ours (full)}               & {0.88} & \textbf{0.58} & {0.21}
                                   & \textbf{0.69} & \textbf{0.43} & {0.37}
                                   & \textbf{0.55} & \textbf{0.31} & {0.40} \\
\bottomrule
\end{tabularx}
\end{table*}

\section{Experiments}
\subsection{Simulation Setup}
\textbf{Dataset and Simulator:} 
We evaluate our framework on the Matterport3D (MP3D) dataset~\cite{ref35}, with 80 scenes for training and 10 for testing. To enable full 3D quadrotor navigation, we adopt the VisFly simulator \cite{ref36}, which augments the original Habitat simulator \cite{ref37} with 3D collision detection and a continuous 3D action space.

\textbf{Episode Settings:} 
The quadrotor’s starting height is randomly sampled between $0.5$ m and $2$ m above the floor. 
Goal images are filtered following \cite{ref5} to ensure that the target object is fully visible and not occluded. Episodes are categorized into three difficulty levels based on the Euclidean distance between the start and goal positions: Easy ($1.5 \sim 4.5$ m), Medium ($4.5 \sim 7$ m), and Hard ($>7$ m).

\textbf{Evaluation Metrics:} 
We adopt Success Rate (SR) and Success Weighted by Path Length (SPL) as evaluation metrics. An episode is successful if the quadrotor reaches within $1.5$ m of the goal and the target is detected in the observation. SPL is defined as
\begin{equation}
\text{SPL} = \frac{1}{N} \sum_{i=1}^{N} S_i \frac{L_i}{\max(P_i, L_i)},
\end{equation}
where $S_i$ is the success indicator, $L_i$ is the length of the planner-generated optimal path, and $P_i$ is the executed path length. The maximum step limit is $150$.
In addition, we report the collision failure rate (CFR), defined as the percentage of failed episodes that terminate due to a collision with obstacles, as it directly reflects the safety of the navigation policy and its ability to generate collision-free trajectories.

\textbf{Implementation Details:} 
RGBD images are rendered at a resolution of $256 \times 256$ with a $90^{\circ} \times 90^{\circ}$ FOV. 
A lightweight Siamese classifier is designed following \cite{ref12}. It issues a stop signal once the target is confidently detected. The object detection is performed using YOLOE-V8s without fine-tuning. 
The graph encoder consists of $L=2$ layers with 128-dimensional hidden features and four attention heads, 
while the pointer-network decoder is implemented as a two-layer MLP with hidden size $128$. 
For frontier generation, the occupancy grid resolution is set to $0.1\,\mathrm{m}$, 
and candidate viewpoints are sampled every $0.5\,\mathrm{m}$ along each frontier cluster with a yaw step of $30^{\circ}$. 
The viewpoint scoring weights are chosen as 
$\lambda_{\mathrm{IG}}=1.0,\ \lambda_{\mathrm{cov}}=1.0,\ \lambda_{\mathrm{yaw}}=0.3,\ \lambda_{\mathrm{path}}=0.1$. 
For reward shaping, we set $\omega_d = 1.0$ and $\omega_s=0.01$. 
The velocity limit for trajectory generation is $2\,\mathrm{m/s}$. 
Training is conducted on a single NVIDIA RTX A6000 GPU. 
The visualization of a successful episode is shown in Fig. 2.

\begin{figure}[!t]
\centering
\includegraphics[width=3.5in]{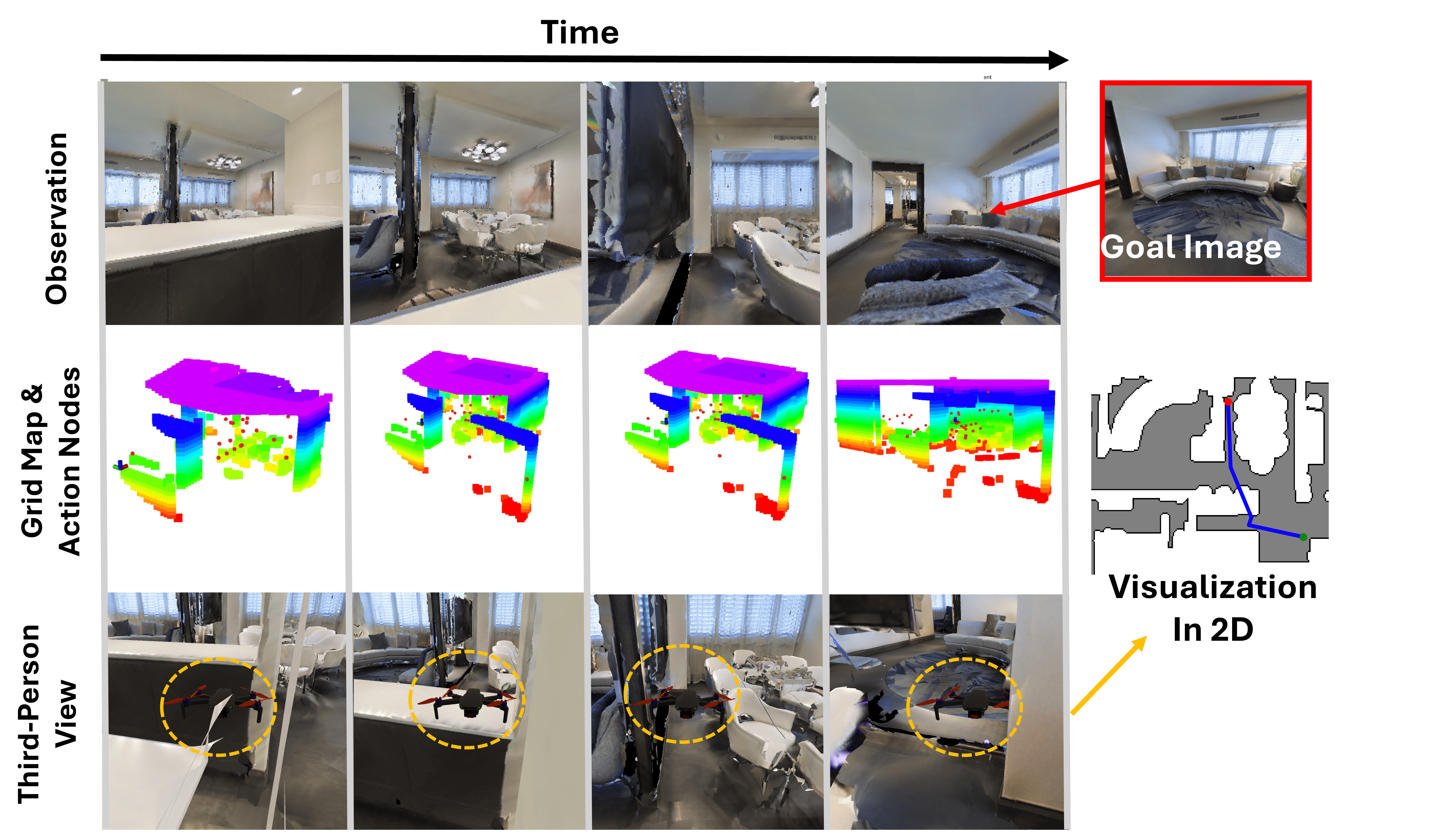}
\captionsetup{font=small}
\caption{Visualization of the simulated quadrotor InstanceImageNav task. The top row shows the quadrotor’s RGB observations, the middle row displays the online-built 3D occupancy grid with red spheres marking the action nodes, and the bottom row provides the third-person view rendered in the VisFly simulator. The rightmost column shows the goal image and the final top-down 2D trajectory map.}
\label{fig_2}
\end{figure}

\begin{table*}[t]
\renewcommand{\arraystretch}{1.1}
\centering
\caption{Ablation study results.}
\small
\label{tab:ablation}
\begin{tabularx}{0.95\textwidth}{>{\raggedright\arraybackslash}p{4.2cm}*{9}{>{\centering\arraybackslash}X}}
\toprule
\multirow{2}{*}{\textbf{Method}} 
 & \multicolumn{3}{c}{Easy} & \multicolumn{3}{c}{Medium} & \multicolumn{3}{c}{Hard} \\
\cmidrule(lr){2-10}
 & SR & SPL & CFR & SR & SPL & CFR & SR & SPL & CFR \\
\midrule
Ours (full)                    &\textbf{0.88} &\textbf{0.58} &0.21  &\textbf{0.69} &\textbf{0.43} &0.37  &\textbf{0.55} &\textbf{0.31} &0.40 \\
w/o Obj-to-Obs Feature Fusion  &0.54 &0.43 &0.18  &0.57 &0.38 &0.29  &0.31 &0.19 &0.35 \\
YOLOE $\rightarrow$ Mask R-CNN &0.59 &0.33 &0.15  &0.46 &0.30 &0.23  &0.24 &0.15 &0.35 \\
w/o Shortcut Nodes             &0.77 &0.46 &0.28  &0.49 &0.27 &0.26  &0.51 &0.23 &0.38  \\
w/o Viewpoint Generation       &0.52 &0.33 &0.35  &0.39 &0.26 &0.49  &0.22 &0.17 &0.67  \\
w/o Trajectory Planner         &0.37 &0.15 &0.60  &0.26 &0.18 &0.59  &0.13 &0.08 &0.89  \\
\bottomrule
\end{tabularx}
\end{table*}

\subsection{Comparative Study}

We compare our method with several representative approaches covering both end-to-end and modular navigation paradigms.

\par\textbf{OVRL-v2-IIN \cite{ref14}:} 
An end-to-end navigation policy that employs a ViT+LSTM architecture together with self-supervised visual pre-training.

\par\textbf{FUEL \cite{ref19}:} 
A classical frontier-based exploration framework that selects local goals by minimizing distance cost without explicitly considering the target object.

\par\textbf{Mod-IIN \cite{ref42}:} 
A modular pipeline consisting of exploration, goal re-identification, goal localization, and local navigation stages.

\par\textbf{Modular ImageNav \cite{ref12}:} 
A modular image-goal navigation method that predicts long-term navigation goals from joint embeddings of the goal image and the current map.

\par\textbf{Topo-metric ImageNav \cite{ref9}:} 
A modular framework that constructs a topo-metric graph and selects action nodes based on frontier exploration.

\par\textbf{IEVE \cite{ref5}:} 
An Instance ImageGoal Navigation framework that introduces an Exploration–Verification–Exploitation (IEVE) strategy with instance segmentation and feature matching to verify targets and avoid distractors.

To ensure a fair comparison in the VisFly simulator, we implement the baselines following their original designs while adapting them to the quadrotor InstanceImageNav setting. 
Since most existing methods are developed for planar navigation, We adapt baselines with minimal necessary interface changes to run in VisFly continuous 3D control, while preserving their high-level logic.
For \textbf{OVRL-v2-IIN}, the original discrete action head is replaced with a continuous control interface that outputs linear velocities and yaw rate, allowing the policy to control the quadrotor directly.
For \textbf{FUEL}, we add a goal-image detection module so that the agent can terminate the exploration once the target instance becomes visible.
For \textbf{Mod-IIN}, \textbf{Modular ImageNav}, \textbf{IEVE}, and \textbf{Topo-metric ImageNav}, their original 2D spatial representations are extended to a 3D occupancy grid to support frontier extraction and goal localization in the aerial navigation setting. 
To ensure consistent motion execution across all methods, we use the same trajectory generation module for all modular approaches.

The results in Table~\uppercase\expandafter{\romannumeral1} show that the proposed approach consistently outperforms all baselines on the quadrotor InstanceImageNav task.
\textbf{OVRL-v2-IIN} achieves the lowest performance in this setting. 
One possible reason is that end-to-end policies must simultaneously learn perception, exploration, and control in a continuous 3D action space, which significantly increases the learning difficulty.
\textbf{FUEL} achieves relatively high success rates but obtains lower SPL. 
Since it performs frontier exploration without incorporating the goal image, the resulting trajectories are often longer before the target instance is observed.
\textbf{Mod-IIN} performs less effectively in complex scenarios because it does not maintain an explicit environment memory and therefore relies mainly on locally visible information when selecting frontiers.
\textbf{Modular ImageNav} predicts navigation targets directly from fused image and map features. 
While flexible, this design may produce redundant or suboptimal goals in 3D environments, which can lead to longer trajectories and increased collision risk.
\textbf{IEVE} relies on heuristic mode switching, limiting robustness in complex 3D environments. 
\textbf{Topo-metric ImageNav} achieves the best performance among the baselines. 
However, it directly treats frontiers as action nodes without considering camera viewing constraints. 
In contrast, our method explicitly generates viewpoint-aware action nodes, which improves frontier coverage and target observation under limited FOV sensing.

\subsection{Ablation Study}
As shown in Table~\uppercase\expandafter{\romannumeral2}, we conduct ablation studies to evaluate the contribution of each major component of the proposed framework. 

\textbf{Environment Memory Module:}
We first analyze two variants that weaken the semantic representation stored in the environment memory. 
In the first variant (Row~2), we remove the cross-layer semantic propagation from the object layer to the observation layer, preventing object-level information from influencing action node selection. 
In the second variant (Row~3), we replace the open-vocabulary detector YOLOE with a closed-set Mask R-CNN detector and substitute the CLIP-based semantic embedding with the corresponding class label. 
This modification evaluates the impact of richer semantic representations on navigation performance. 
Both variants lead to noticeable drops in SR and SPL, particularly under the Hard setting. 
Without cross-layer semantic propagation, the policy relies mainly on local visual features when evaluating candidate action nodes. 
Replacing the open-vocabulary detector further reduces the diversity of detectable objects, limiting the system's ability to associate observations with the target instance.

\textbf{Action Node Generation Module:}
We further study the role of viewpoint-aware action node generation. 
First, we remove the shortcut nodes produced through feature matching (Row~4), forcing the agent to rely solely on frontier exploration. 
Second, we disable viewpoint generation and directly use frontier cluster centroids as action nodes (Row~5). 
Removing shortcut nodes delays goal verification and often results in longer trajectories, leading to lower SPL. 
When viewpoint generation is disabled, the quadrotor frequently approaches candidate regions from unfavorable viewing directions, reducing the likelihood of observing the target instance and increasing collision risk near obstacles.

\textbf{Trajectory Planning Module:}
Finally, we evaluate the impact of the trajectory generation module. 
In this variant, the policy network directly outputs velocity and yaw-rate commands without using the trajectory planner. 
This modification results in a significant drop in SR and a clear increase in collision failures. 
The results highlight the importance of separating high-level decision making from low-level motion execution. 
Without the optimization-based planner, the policy must simultaneously learn perception, exploration, and safe flight control in cluttered 3D environments, which substantially increases the learning difficulty.

\begin{figure*}[t] 
    \centering
    \includegraphics[width=\textwidth]{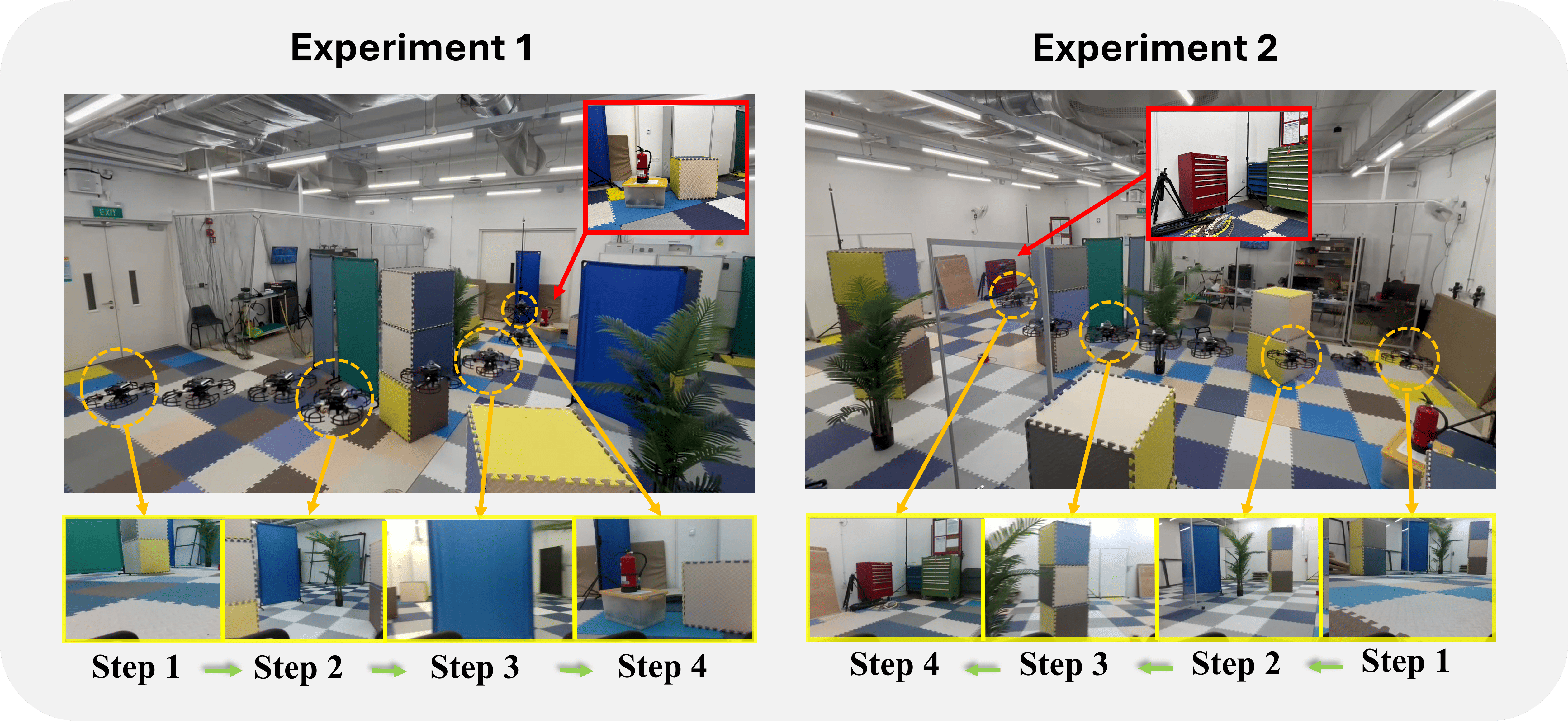}
    \caption{Two representative flight trajectories of the quadrotor performing the InstanceImageNav task in real-world indoor environments. The goal images are shown in red boxes, and the observed RGB frames along the trajectory are shown in yellow boxes. In both episodes, the quadrotor navigates from one end of the room to the other before stopping at the target.}
    \label{fig:scene_example}
\end{figure*}

\subsection{Navigation in the Real World}
We further conduct real-world quadrotor InstanceImageNav experiments to validate the effectiveness of the proposed navigation framework. Our customized quadrotor platform is equipped with a Jetson Orin NX for onboard computation, a Livox MID-360 for localization, a Pixhawk flight controller for trajectory execution, and a forward-facing RealSense D435i for visual perception. Real-time localization is achieved by running the Fast-LIO algorithm \cite{ref43} onboard.
Figure. 3 presents two successful episodes. In each case, the quadrotor begins from an initial pose with its camera oriented away from the target object. The quadrotor then explores the unknown indoor environment, detects the target, and issues a stop command.

\section{Conclusion}
This paper studies the InstanceImageNav problem for quadrotor platforms operating in continuous 3D environments with limited FOV sensing. We show that successful navigation in this setting critically depends on selecting informative viewpoints rather than simply reaching spatial locations.
To address this challenge, we develop a hierarchical navigation framework that combines environment memory, viewpoint-aware action node generation, and trajectory planning. The proposed system enables the quadrotor to reason over accumulated semantic and spatial context while safely navigating toward candidate viewpoints. By separating high-level decision making from low-level motion execution, the framework integrates learning-based policies with reliable optimization-based planning.
Extensive experiments in simulation demonstrate consistent improvements over representative baselines, particularly in complex environments. Real-world quadrotor flights further validate the practicality and robustness of the proposed system.





\balance


\begin{thebibliography}{99}

\bibitem{ref1}
Sakaguchi, T., Taniguchi, A., Hagiwara, Y., Hafi, L. E., Hasegawa, S., and Taniguchi, T. (2024). Real-world Instance-specific Image Goal Navigation for Service Robots: Bridging the Domain Gap with Contrastive Learning. arXiv preprint arXiv:2404.09645.
\bibitem{ref2}
Stokkeland M, Klausen K, Johansen T A. Autonomous visual navigation of unmanned aerial vehicle for wind turbine inspection[C]//2015 International Conference on Unmanned Aircraft Systems (ICUAS). IEEE, 2015: 998-1007.
\bibitem{ref3}
Xiao J, Pisutsin P, Feroskhan M. Collaborative target search with a visual drone swarm: An adaptive curriculum embedded multistage reinforcement learning approach[J]. IEEE Transactions on Neural Networks and Learning Systems, 2023.
\bibitem{ref4}
Krantz, J., Lee, S., Malik, J., Batra, D., and Chaplot, D. S. (2022). Instance-specific image goal navigation: Training embodied agents to find object instances. arXiv preprint arXiv:2211.15876.
\bibitem{ref5}
Lei X, Wang M, Zhou W, et al. Instance-aware Exploration-Verification-Exploitation for Instance ImageGoal Navigation[C]//Proceedings of the IEEE/CVF Conference on Computer Vision and Pattern Recognition. 2024: 16329-16339.
\bibitem{ref6}
Wasserman J, Yadav K, Chowdhary G, et al. Last-mile embodied visual navigation[C]//Conference on Robot Learning. PMLR, 2023: 666-678.
\bibitem{ref7}
Sun X, Chen P, Fan J, et al. FGPrompt: fine-grained goal prompting for image-goal navigation[J]. Advances in Neural Information Processing Systems, 2024, 36.
\bibitem{ref8}
Kim N, Kwon O, Yoo H, et al. Topological semantic graph memory for image-goal navigation[C]//Conference on Robot Learning. PMLR, 2023: 393-402.
\bibitem{ref9}
Ye S, Cui Y, Sha H, et al. RGBD-based Image Goal Navigation with Pose Drift: A Topo-metric Graph based Approach[C]//2024 IEEE International Conference on Robotics and Automation (ICRA). IEEE, 2024: 18391-18397.
\bibitem{ref10}
Yan Z, Huang R, He L, et al. SIGN: Safety-Aware Image-Goal Navigation for Autonomous Drones via Reinforcement Learning[J]. arXiv preprint arXiv:2508.12394, 2025.
\bibitem{ref11}
Gervet T, Chintala S, Batra D, et al. Navigating to objects in the real world[J]. Science Robotics, 2023, 8(79): eadf6991.
\bibitem{ref12}
Wu Q, Wang J, Liang J, et al. Image-goal navigation in complex environments via modular learning[J]. IEEE Robotics and Automation Letters, 2022, 7(3): 6902-6909.

\bibitem{ref13}
Chaplot D S, Salakhutdinov R, Gupta A, et al. Neural topological slam for visual navigation[C]//Proceedings of the IEEE/CVF conference on computer vision and pattern recognition. 2020: 12875-12884.
\bibitem{ref14}
Yadav K, Majumdar A, Ramrakhya R, et al. Ovrl-v2: A simple state-of-art baseline for imagenav and objectnav[J]. arXiv preprint arXiv:2303.07798, 2023.
\bibitem{ref15}
Chen, K., De Vicente, J. P., Sepulveda, G., Xia, F., Soto, A., Vázquez, M., and Savarese, S. (2019). A behavioral approach to visual navigation with graph localization networks. arXiv preprint arXiv:1903.00445.
\bibitem{ref16}
Shah D, Eysenbach B, Kahn G, et al. Ving: Learning open-world navigation with visual goals[C]//2021 IEEE International Conference on Robotics and Automation (ICRA). IEEE, 2021: 13215-13222.
\bibitem{ref17}
Suomela L, Kalliola J, Edelman H, et al. Placenav: Topological navigation through place recognition[C]//2024 IEEE International Conference on Robotics and Automation (ICRA). IEEE, 2024: 5205-5213.
\bibitem{ref18}
Yamauchi B. Frontier-based exploration using multiple robots[C]//Proceedings of the second international conference on Autonomous agents. 1998: 47-53.
\bibitem{ref19}
Zhou B, Zhang Y, Chen X, et al. Fuel: Fast uav exploration using incremental frontier structure and hierarchical planning[J]. IEEE Robotics and Automation Letters, 2021, 6(2): 779-786.
\bibitem{ref20}
Maksymets O, Cartillier V, Gokaslan A, et al. Thda: Treasure hunt data augmentation for semantic navigation[C]//Proceedings of the IEEE/CVF International Conference on Computer Vision. 2021: 15374-15383.
\bibitem{ref21}
Bono G, Antsfeld L, Chidlovskii B, et al. End-to-end (instance)-image goal navigation through correspondence as an emergent phenomenon[J]. arXiv preprint arXiv:2309.16634, 2023.
\bibitem{ref22}
Mezghan L, Sukhbaatar S, Lavril T, et al. Memory-augmented reinforcement learning for image-goal navigation[C]//2022 IEEE/RSJ International Conference on Intelligent Robots and Systems (IROS). IEEE, 2022: 3316-3323.
\bibitem{ref23}
Ramakrishnan S K, Chaplot D S, Al-Halah Z, et al. Poni: Potential functions for objectgoal navigation with interaction-free learning[C]//Proceedings of the IEEE/CVF Conference on Computer Vision and Pattern Recognition. 2022: 18890-18900.
\bibitem{ref24}
Chaplot, D. S., Gandhi, D., Gupta, S., Gupta, A., and Salakhutdinov, R. (2020). Learning to explore using active neural slam. arXiv preprint arXiv:2004.05155.
\bibitem{ref25}
Chen, T., Gupta, S., and Gupta, A. (2019). Learning exploration policies for navigation. arXiv preprint arXiv:1903.01959.
\bibitem{ref26}
Ravichandran Z, Peng L, Hughes N, et al. Hierarchical representations and explicit memory: Learning effective navigation policies on 3d scene graphs using graph neural networks[C]//2022 International Conference on Robotics and Automation (ICRA). IEEE, 2022: 9272-9279.
\bibitem{ref27}
An D, Wang H, Wang W, et al. Etpnav: Evolving topological planning for vision-language navigation in continuous environments[J]. IEEE Transactions on Pattern Analysis and Machine Intelligence, 2024.
\bibitem{ref28}
Werby A, Huang C, Büchner M, et al. Hierarchical Open-Vocabulary 3D Scene Graphs for Language-Grounded Robot Navigation[C]//First Workshop on Vision-Language Models for Navigation and Manipulation at ICRA 2024. 2024.
\bibitem{ref29}
Hart P E, Nilsson N J, Raphael B. A formal basis for the heuristic determination of minimum cost paths[J]. IEEE transactions on Systems Science and Cybernetics, 1968, 4(2): 100-107.
\bibitem{ref30}
Potje G, Cadar F, Araujo A, et al. XFeat: Accelerated Features for Lightweight Image Matching[C]//Proceedings of the IEEE/CVF Conference on Computer Vision and Pattern Recognition. 2024: 2682-2691.
\bibitem{ref31}
Lindenberger P, Sarlin P E, Pollefeys M. Lightglue: Local feature matching at light speed[C]//Proceedings of the IEEE/CVF International Conference on Computer Vision. 2023: 17627-17638.
\bibitem{ref32}
Schulman J, Wolski F, Dhariwal P, et al. Proximal policy optimization algorithms[J]. arXiv preprint arXiv:1707.06347, 2017.
\bibitem{ref33}
Wang Z, Chen J, Chen H. EGAT: Edge-featured graph attention network[C]//Artificial Neural Networks and Machine Learning–ICANN 2021: 30th International Conference on Artificial Neural Networks, Bratislava, Slovakia, September 14–17, 2021, Proceedings, Part I 30. Springer International Publishing, 2021: 253-264.
\bibitem{ref34}
Vinyals O, Fortunato M, Jaitly N. Pointer networks[J]. Advances in neural information processing systems, 2015, 28.
\bibitem{ref35}
Chang A, Dai A, Funkhouser T, et al. Matterport3d: Learning from rgb-d data in indoor environments[J]. arXiv preprint arXiv:1709.06158, 2017.
\bibitem{ref36}
Li, F., Sun, F., Zhang, T., and Zou, D. (2024). VisFly: An Efficient and Versatile Simulator for Training Vision-based Flight. arXiv preprint arXiv:2407.14783.
\bibitem{ref37}
Savva M, Kadian A, Maksymets O, et al. Habitat: A platform for embodied ai research[C]//Proceedings of the IEEE/CVF international conference on computer vision. 2019: 9339-9347.
\bibitem{ref38}
He K, Gkioxari G, Dollár P, et al. Mask r-cnn[C]//Proceedings of the IEEE international conference on computer vision. 2017: 2961-2969.
\bibitem{ref39}
J. Li, P. Zhou, C. Xiong, and S. C. Hoi. Prototypical Contrastive Learning of Unsupervised Representations. International Conference on Learning Representations (ICLR), 2021.
\bibitem{ref40}
Wang A, Liu L, Chen H, et al. Yoloe: Real-time seeing anything[J]. arXiv preprint arXiv:2503.07465, 2025.
\bibitem{ref41}
Radford A, Kim J W, Hallacy C, et al. Learning transferable visual models from natural language supervision[C]//International conference on machine learning. PmLR, 2021: 8748-8763.
\bibitem{ref42}
Krantz J, Gervet T, Yadav K, et al. Navigating to objects specified by images[C]//Proceedings of the IEEE/CVF International Conference on Computer Vision. 2023: 10916-10925.
\bibitem{ref43}
Xu W, Zhang F. Fast-lio: A fast, robust lidar-inertial odometry package by tightly-coupled iterated kalman filter[J]. IEEE Robotics and Automation Letters, 2021, 6(2): 3317-3324.
\bibitem{ref44}
Xiao J, Zhang R, Zhang Y, et al. Vision-based learning for drones: A survey[J]. IEEE Transactions on Neural Networks and Learning Systems, 2025.

\end{thebibliography}
\end{document}